\documentclass{article}
\usepackage{arxiv}

\usepackage{hyperref}
\usepackage{url}
\usepackage{listings}
\usepackage{graphicx}
\usepackage{xcolor}
\usepackage{amsfonts}
\usepackage{amsmath}
\RequirePackage{natbib}

\title{Adversarial Robustness of \\ Flow-Based Generative Models}

\author{ Phillip Pope\thanks{First two authors contributed equally} \\ \texttt{pepope@cs.umd.edu} 
    \And Yogesh Balaji$^*$ \\ \texttt{yogesh@cs.umd.edu}  
    \And  Soheil Feizi \\   \texttt{sfeizi@cs.umd.edu}
}
\date{
\vspace{-4ex}
Department of Computer Science\\
 University of Maryland, College Park\\
 College Park, MD 20740, USA \\
}

\renewcommand{\headeright}{}
\renewcommand{\undertitle}{}
 
\begin{document}
\maketitle

\newcommand{\bx}{\mathbf{x}}
\newcommand{\tx}{\tilde{\mathbf{x}}}
\newcommand{\cN}{\mathcal{N}}
\newcommand{\bE}{\mathbb{E}}
\newcommand{\aI}{\alpha I}

\newtheorem{theorem}{Theorem}[section]
\newtheorem{corollary}{Corollary}[theorem]
\newtheorem{lemma}[theorem]{Lemma}

\begin{abstract}
Flow-based generative models leverage invertible generator functions to fit a distribution to the training data using maximum likelihood. Despite their use in several application domains, robustness of these models to adversarial attacks has hardly been explored. In this paper, we study adversarial robustness of flow-based generative models both theoretically (for some simple models) and empirically (for more complex ones). First, we consider a linear flow-based generative model and compute optimal sample-specific and universal adversarial perturbations that maximally decrease the likelihood scores. Using this result, we study the robustness of the well-known \textit{adversarial training} procedure, where we characterize the fundamental trade-off between model robustness and accuracy. Next, we empirically study the robustness of two prominent deep, non-linear, flow-based generative models, namely GLOW and RealNVP. We design two types of adversarial attacks; one that minimizes the likelihood scores of in-distribution samples, while the other that maximizes the likelihood scores of out-of-distribution ones. We find that GLOW and RealNVP are extremely sensitive to both types of attacks. Finally, using a hybrid adversarial training procedure, we significantly boost the robustness of these generative models.

\end{abstract}

\section{Introduction}

The promise of modern deep generative models is to learn data distributions with sufficiently high fidelity, allowing simulation of realistic samples. Some applications include photo-realistic image generation, audio synthesis, and image to text generation \citep{pmlr-v48-reed16, Ledig_2017_CVPR, DBLP:journals/corr/OordDZSVGKSK16}. Generative Adversarial Networks (GANs)\citep{NIPS2014_5423} have become a popular choice in modern generative modeling, often obtaining the state-of-the-art results in image and video synthesis \citep{DBLP:conf/iclr/KarrasALL18}. While GANs can synthesize samples from a data distribution, their inability to compute sample likelihoods limits their usage in statistical inference tasks~\citep{balaji2019entropicgan}. 

\begin{figure*}
\begin{center}
{\includegraphics[width=\textwidth]{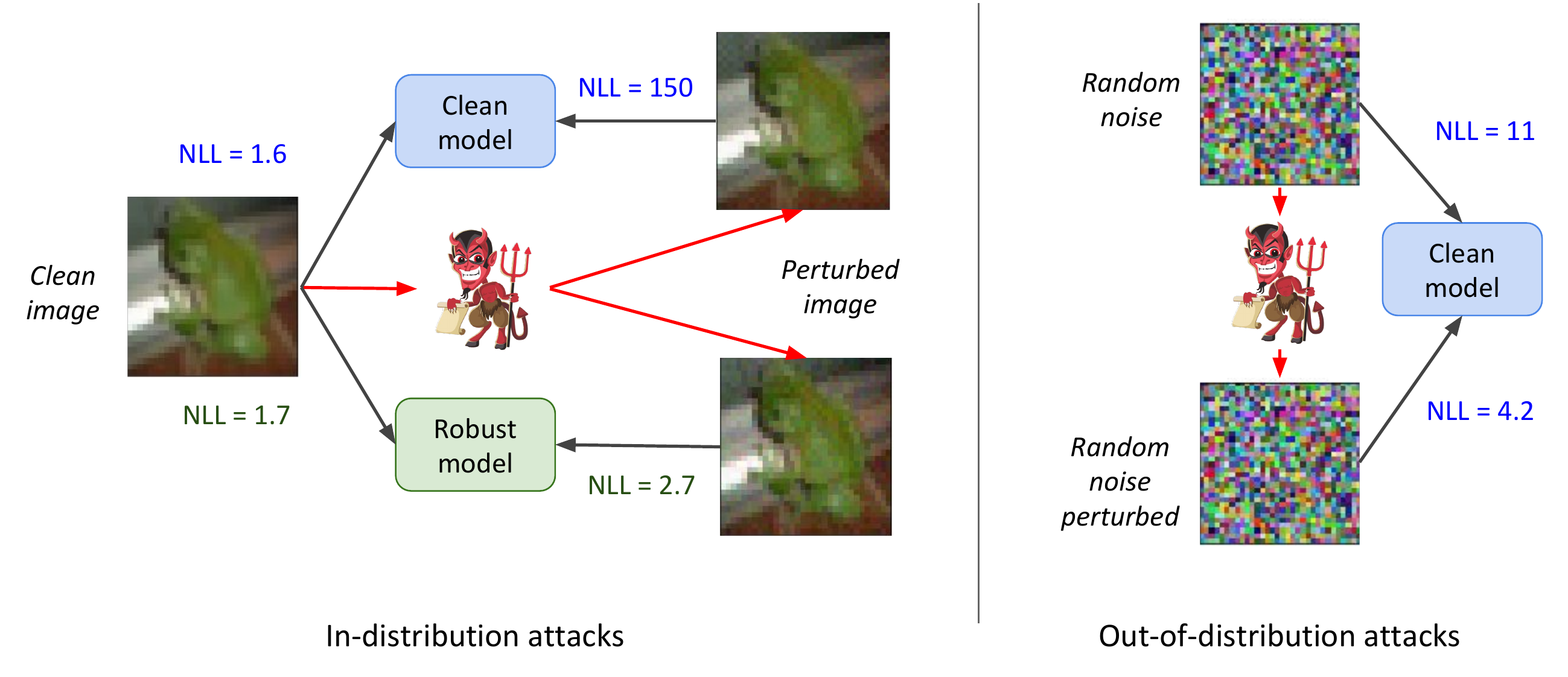}}
\end{center}
\caption{Sensitivity of flow-based generative models to adversarial attacks. A low NLL indicates a high likelihood score. The figure on the left panel shows \textit{in-distribution} attacks: a frog image which is assigned a NLL score of $1.6$ by GLOW model trained on CIFAR-10 (clean model) gives a high NLL of $150$ when perturbed adversarially ($\epsilon=8$ in $\ell_{\infty}$ norm). Our robust model significantly improves the robustness: NLL does not change much after the attack, while the score on the unperturbed sample is similar to the one obtained by the clean model. The panel on the right shows \textit{out-of-distribution} attacks where a noise image is assigned a low NLL of $4.2$.}
\label{fig:title}
\end{figure*}

Likelihood-based models, on the other hand, explicitly fit a generative model to the data using a maximum likelihood optimization, enabling exact or approximate evaluations of sample likelihoods at the test time. Some popular choices include auto-regressive models \citep{NIPS2016_6527}, ~\citep{pmlr-v48-oord16}, Variational Auto-encoders~\citep{DBLP:journals/corr/abs-1906-02691}, and methods based on normalizing flow~\citep{pmlr-v37-rezende15}. Notably, flow-based models~\citep{NIPS2018_8224, DBLP:conf/iclr/DinhSB17} leverage invertible generator functions to learn a bijective mapping between latent space and the data distribution, enabling an exact sample likelihood computation.

The focus of this paper is to perform a comprehensive study of robustness of likelihood-based generative models to adversarial perturbations of their inputs. While there has been progress on adversarial robustness of classification problems~\citep{DBLP:conf/iclr/MadryMSTV18}, robustness of likelihood models has not been explored in the literature. Performing such a sensitivity analysis is crucial for reliable deployment, especially in safety-critical applications. For instance, one application where likelihood estimation is crucial is unsupervised anomaly detection in medical imaging, where out-of-distribution samples can be detected using likelihood scores. Adversarial attacks on such systems can lead to false diagnosis, potentially bearing life-threatening consequences. 


First, we present a theoretical analysis of the sensitivity of linear generative models that fit a Gaussian distribution to the data (since the latent variable is often a Gaussian distribution itself). Under this setting, we compute the optimal sample-specific and universal norm-bounded input perturbations to maximally decrease the likelihood scores. We then analyze the effectiveness of one of the most successful defense mechanism against adversarial attacks, namely \textit{adversarial training} \citep{DBLP:conf/iclr/MadryMSTV18} where adversarially perturbed samples are recursively used in re-training the model. We show that adversarial training can provably defend against norm-bounded adversarial attacks. However, this comes at a cost of decrease in clean likelihood scores. This naturally gives rise to a fundamental trade-off between model performance and robustness. 

Next, we empirically show the existence of adversarial attacks on two popular deep flow-based generative models: GLOW~\citep{NIPS2018_8224} and RealNVP~\citep{DBLP:conf/iclr/DinhSB17}. A measure of likelihood, by definition, should assign low scores to out-of-distribution samples and high scores to in-distribution ones. The existence of adversarial attacks breaks and contradicts this intuition: We show that we can construct samples that look like normal (in-distribution) data to a human eye, yet the model assigns to them low likelihood scores, or equivalently, high negative $\log$ likelihood (NLL) scores. Similarly, we show the existence of out-of-distribution samples that are assigned low NLL scores. One such example is shown in Figure \ref{fig:title}, where a sample from CIFAR-10 dataset when adversarially perturbed has high NLL score, and a random adversarially perturbed image (with uniform pixel intensities) has low NLL score. This observation raises serious doubts about the reliability of likelihood scores obtained through standard flow-based models.

To make these models robust, we investigate the effect of the popular \textit{adversarial training} mechanism. We show that adversarial training empirically improves robustness, however, this comes at a cost of decrease in likelihood scores on unperturbed test samples compared to the baseline model. To mitigate this effect, we propose a novel variant of adversarial training, called the {\it hybrid adversarial training}, where the negative log likelihoods of both natural and perturbed samples are minimized during training. We show that hybrid adversarial training obtains increased robustness on adversarial examples, while simultaneously maintaining high likelihood scores on clean test samples.
 
In summary, our contributions are as follows: 

\begin{itemize}
\setlength\itemsep{0.5em}
\item We theoretically analyze the robustness of linear generative models, and show that adversarial training provably learns robust models. We also characterize the fundamental trade-off between model robustness and performance.
\item We demonstrate the existence of \textit{in-distribution} and \textit{out-of-distribution} attacks on flow-based likelihood models.
\item We propose a novel variant of adversarial training, called the \textit{hybrid adversarial training}, that can learn robust flow-based models while maintaining high likelihood scores on unperturbed samples.
\end{itemize}

\section{Background}
\subsection{Flow-based Generative models}




Although generative modeling is largely dominated by generative adversarial networks (GANs), one major short-coming of GANs is its inability to compute sample likelihoods. 
\textit{Flow-based} generative models solve this issue by designing an invertible transformation $f: \mathbb{R}^{D} \rightarrow \mathbb{R}^D$ between latent distribution $p_{\mathbf{z}}(\mathbf{z})$ and the generated distribution $p_{\mathbf{x}}(\mathbf{x})$. $p_{\mathbf{z}}(\mathbf{z})$ is often assumed to be a normal distribution. Given a random variable $\mathbf{z} \sim p_{\mathbf{z}}(\mathbf{z})$, we can use \textit{change of variables} to write the log density of a sample $\mathbf{x}$ such that $\mathbf{x}=f(\mathbf{z})$ as \citep{DBLP:journals/corr/DinhKB14, DBLP:conf/iclr/DinhSB17, DBLP:conf/iclr/GrathwohlCBSD19, NIPS2018_8224}:
\begin{align*}
  \log p_\mathbf{x}(\mathbf{x}) = \log p_{\mathbf{z}}(\mathbf{z}) - \log \det \left| \frac{\partial f(\mathbf{z})}{\partial \mathbf{z}} \right| 
\end{align*}
Since $f$ is invertible, inference can be performed as $\mathbf{z} = f^{-1}(\mathbf{x})$. The transformation $f$ is typically modeled as the composition of $K$ invertible maps, $f = f_1 \circ f_2 \circ \cdots \circ f_K$, also called \textit{normalizing flows}~\citep{pmlr-v37-rezende15}. The special structure used in each $f_i$'s allows an efficient computation of the determinant.

Prominent examples of flow-based generative models include NICE \citep{DBLP:journals/corr/DinhKB14}, RealNVP \citep{DBLP:conf/iclr/DinhSB17}, GLOW \citep{NIPS2018_8224}, and FFJORD \citep{DBLP:conf/iclr/GrathwohlCBSD19}, each using a particular choice of $f_k$ . For instance, RealNVP \citep{DBLP:conf/iclr/DinhSB17} is designed with \textit{affine coupling layers}, essentially an invertible scale transformation, while GLOW uses \textit{invertible $1\times 1$ convolutions} \citep{NIPS2018_8224}, which utilizes learned permutations.

\subsection{Adversarial Attacks and Robustness}\label{sec:background_adv_rob}
\label{sec:adv-attacks-and-robustness}

In context of classification, adversarial examples are subtle input perturbations that changes a model prediction. These perturbations are small in the sense of a suitable norm and imperceptible to humans. The existence of such examples raises serious concern for the deployment of machine learning models in safety-critical applications 

Let $\mathcal{D} = \{ (\mathbf{x}, y)\}_{i=1}^N$ be a collection of labeled input instances, $\theta$ be the parameters of a classifier with loss function $L_{cls}$ (e.g. cross-entropy). For a given $\mathbf{x}$, let $S$ be the set of all $\ell_{p}$ norm-bounded perturbations around $\bx$, i.e., $\| \delta \|_{p} < \epsilon$, where $\epsilon$ is a constant, also called the perturbation radius. The perturbed sample is then given by $\bx^{adv} = \bx + \delta$. Standard methods of crafting adversarial examples for classification include the Fast Gradient Sign Method (FGSM) \citep{DBLP:journals/corr/GoodfellowSS14}
\begin{align}
\bx^{adv} = \mathbf{x} + \alpha \, \text{sign} (\nabla_{\mathbf{x}} L_{cls}(\theta,\mathbf{x},y)) \nonumber
\end{align}
and its variant Projected Gradient Descent (PGD) ~\citep{DBLP:conf/iclr/KurakinGB17}
\begin{align}
&\mathbf{x}^{(t+1)} = \text{Proj}_{\mathbf{x}+S}\left(\mathbf{x}^{(t)} + \alpha \, \text{sign}(\nabla_{\mathbf{x}^{(t)}} L_{cls}(\theta, \mathbf{x}^{(t)},y))\right)\nonumber \\
&\bx^{adv} = \bx^{(m)}\nonumber
\end{align}
which is essentially a recursive application of FGSM for $m$ steps, while constraining the perturbation to stay within the feasible region in each step. The above attacks have been shown effective on a variety of datasets \citep{7958570, Xie_2019_CVPR}. For defense against such attacks, \citet{DBLP:conf/iclr/MadryMSTV18} proposed \textit{adversarial training}, a procedure where the parameters $\theta$ of the model is optimized using the following minimax objective 
\begin{align}
\min_{\theta} \mathbb{E}_{(\mathbf{x},y) \sim \mathcal{D}} \left[ \max_{\delta \in S} L_{cls}(\theta, \mathbf{x} + \delta, y) \right] \nonumber
\end{align}
Intuitively, this amounts to training the model on adversarial examples instead of the unperturbed ones. Adversarial training remains to be one of the successful defense mechanisms to date.


\subsection{Robustness of generative models}
To the best of our knowledge, no prior work exists on adversarial robustness of likelihood-based generative models. In \citet{8424630}, attacks on encoder-based generative models (VAEs and VAE-GANs) are constructed to adversarially manipulate reconstruction and latent-space tasks. Adversarial attacks on generative classifiers are explored in \citet{DBLP:journals/corr/abs-1906-01171}, where they show that even near optimal conditional generative models are susceptible to adversarial attacks when used for classfication tasks. \citet{DBLP:conf/iclr/NalisnickMTGL19} discuss some non-intuitive properties of flow-based generative models, studying GLOW in particular. They empirically observe that on a variety of common datasets, GLOW models assign \textit{lower} likelihoods to in-distribution data than out-of-distribution.



\section{Adversarial robustness of linear flow-based generative models}\label{sec:theory}

We begin with an analysis of adversarial robustness of linear flow-based generative models. Since the latent variable $Z$ has a normal distribution and the transformation between $Z$ and $X$ is considered to be affine, $X$ will have a distribution in the form of $X \sim \cN(\mu, K)$ where $\mu$ is the mean vector and $K$ is the covariance matrix. Given $N$ samples as the input dataset $D = \{ \bx_i \}_{i=1}^{N}$, we are interested in estimating the mean and covariance matrices of the generative distribution. This problem can be solved using maximum-likelihood estimation, in which we find model parameters that maximize the likelihood of the input dataset $D$. We know that log-likelihood of a test point $\bx$ under the Gaussian distribution $X$ can be written as 
\begin{align*}
    L(\bx) = C - \frac{1}{2}\log(|K|) - \frac{(\bx - \mu)^{T}K^{-1}(\bx - \mu)}{2}
\end{align*}
where $C=n\log(2\pi)/2$. It is a well-known result that maximizing the log-likelihood of the dataset $D$ results in the following estimators for mean and covariance:
\begin{align*}
    \hat{\mu} &= \frac{\sum_{i} \bx_i}{N} \\
    \hat{K} &= \frac{1}{N} \sum_{i=1}^{N} \frac{(\bx_i - \hat{\mu})(\bx_i - \hat{\mu})^T}{N},
\end{align*}
referred to as the sample mean and sample covariance, respectively.
\subsection{Adversarial attacks}
In this section, we aim to find a norm-bounded adversarial perturbation $\delta$ that maximally decreases the likelihood score of a test point $\bx$. We consider adversarial perturbations with a bounded $\ell_{2}$ norm (i.e., $\| \delta \|_2 < \epsilon$). Please note that while we use $\ell_2$ norm here, $\ell_{\infty}$ perturbation norm bounds are used to construct attacks for non-linear flow-models trained on vision datasets (Section.~\ref{experiments}). We would like to point out that both $\ell_2$ and $\ell_{\infty}$ are commonly used settings to study adversarial robustness~\citep{DBLP:conf/iclr/MadryMSTV18}. The perturbation $\delta$ can be found by solving the following optimization problem
\begin{align}\label{eq:pert_gen}
    \min_{\delta} ~~ & C - \frac{1}{2}\log(|K|) - \frac{(\bx - \mu + \delta)^{T}K^{-1}(\bx - \mu + \delta)}{2} \\
    \text{s.t. } & \| \delta \|_{2} < \epsilon \nonumber
\end{align}

\begin{figure}[h]
\begin{center}
{\includegraphics[height=5cm]{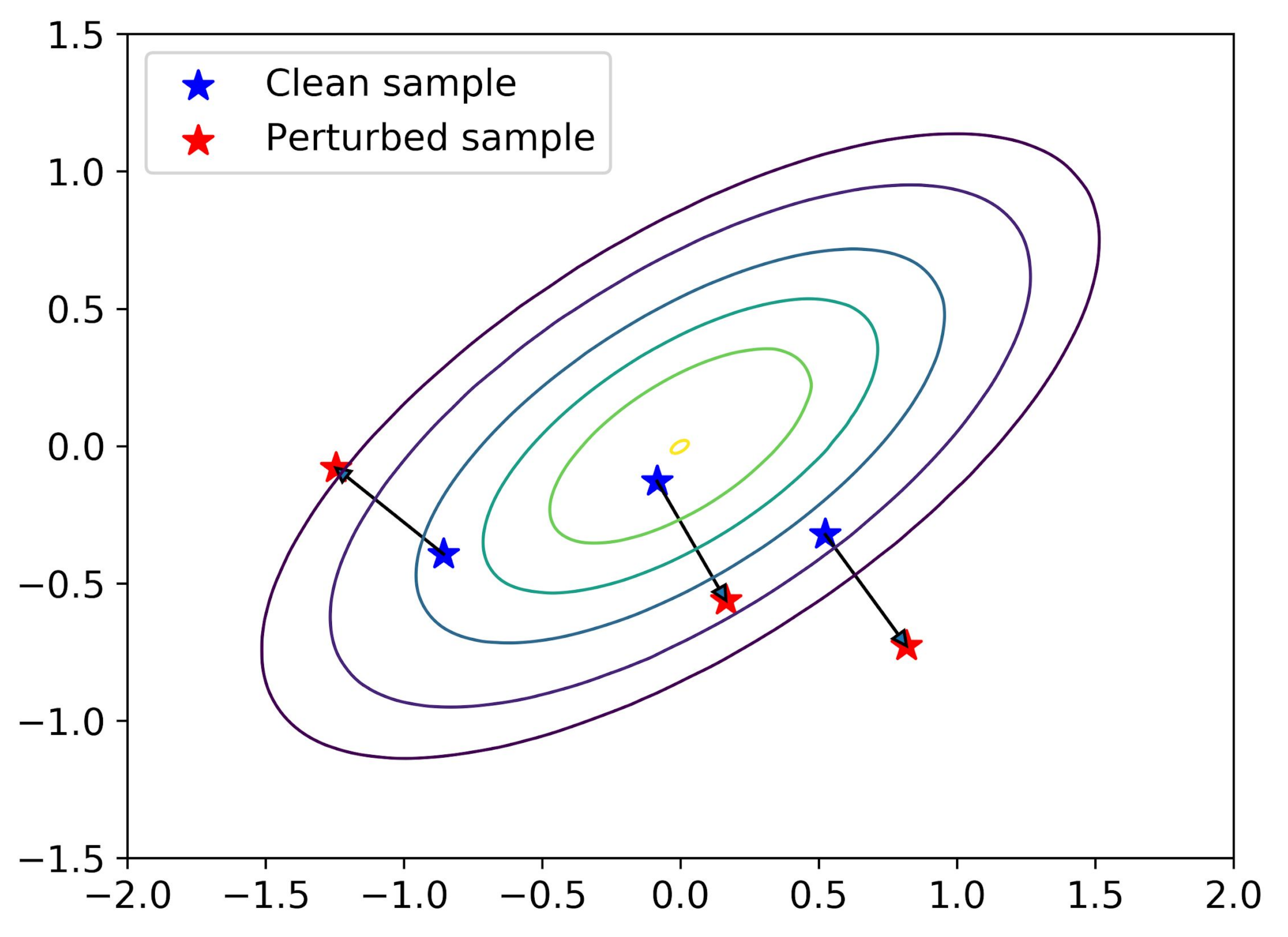}}
\end{center}
\caption{Examples of optimal adversarial perturbations with bounded $\ell_2$ norm for a 2-dimensional Gaussian case.}

\label{fig:gauss-perturb-2d}
\end{figure}

\begin{theorem}\label{thm:pert_gen}
  Let $L(\bx)$ denote the likelihood function of an input sample $\bx$ under a Gaussian distribution $\cN(\mu, K)$. Let $K = U \Lambda U^T$ be the eigen-decomposition of the covariance matrix $K$. Let $c = [c_1, c_2, \hdots, c_n] = U^T K$, and $\Lambda = diag([\lambda_1, \hdots \lambda_n])$. Let $\eta$ be a solution of the set of equations
    \begin{align*}
    \sum_{i} \frac{c_i^{2}}{(1 - 2\eta \lambda_i)^{2}} &= \epsilon^2 \\
    2\eta\lambda_i - 1 &\geq 0 \quad \forall i
    \end{align*}
  Then, the optimal additive perturbation $\delta$ with norm bound $\| \delta \|_{2} < \epsilon$ that maximally decreases the likelihood score of sample $\bx$ is given by
  \begin{align}\label{eq:pert_soln}
      \delta^* = (K^{-1} - 2\eta I)^{-1} K^{-T} (\mu - \bx)
  \end{align}
\end{theorem}
The proof of Theorem~\ref{thm:pert_gen} is given in supplementary material. The above theorem gives a solution for the optimal adversarial perturbation with a bounded $\ell_2$ norm for the linear flow-based generative models (solution to Eq.~\eqref{eq:pert_gen}). Example optimal adversarial perturbations calculated for a 2-dimensional Gaussian distribution are visualized in Figure \ref{fig:gauss-perturb-2d}. Next, we show two special cases of this result.

\subsubsection{Special case: Spherical covariance matrix}
In this case, the covariance matrix is $K=\sigma^{2}I$. In this setup, the solution for the adversarial perturbation problem \eqref{eq:pert_soln} simplifies to:
\begin{align*}
    \delta = \frac{1}{1 - 2\eta\lambda} (\mu - \bx)
\end{align*}
Similarly, the condition for $\eta$ simplifies to:
\begin{align*}
    1 - 2\eta\lambda = \frac{\| \mu - \bx\|}{\epsilon}
\end{align*}
Thus, the optimal adversarial perturbation $\delta$ in this case is given by
\begin{align}\label{eq:pert_soln_linear}
    \delta^* =  \frac{\epsilon}{\| \bx - \mu \|}(\bx - \mu)
\end{align}

\subsubsection{Special case: $\bx = \mu$}
In this case, the optimization \ref{eq:pert_gen} simplifies to:
\begin{align*}
    \min_{\delta} ~~ & C - \frac{1}{2}\log(|K|) - \frac{ \delta^{T}K^{-1}\delta}{2} \\
    \text{s.t. } & \| \delta \|_{2} < \epsilon
\end{align*}
This is a Rayleigh quotient problem, the solution for which is the maximum eigenvalue of $K^{-1}$. I.e.,
\begin{align*}
    \delta^* = \epsilon \, u_{min}(K)
\end{align*}
where $u_{min}$ is the eigenvector of $K$ corresponding to the minimum eigenvalue. Intuitively, minimum eigen-vector of the covariance matrix $K$ is the direction in which the data varies the least, a perturbation along this direction induces a maximal drop in likelihood.

\subsection{Defense against adversarial attacks}
One of the most successful defense strategies against adversarial attacks is adversarial training (Section.~\ref{sec:background_adv_rob}), in which models are recursively trained on adversarially perturbed samples instead of clean ones. In this section, we analyze the effect of adversarial training for the likelihood estimation in the spherical Gaussian case. For an input sample $\bx$ under the generative distribution $\cN(\mu, \sigma^2 I)$, the adversarially perturbed sample using \eqref{eq:pert_soln_linear} is given by
\begin{align*}
    \bx^{adv} &= \bx + \frac{\epsilon}{\| \bx - \mu \|} (\bx - \mu)
\end{align*}
We consider the population case where $\bx\sim \cN(\mu, \sigma^2 I)$. Denote $\tx = (\bx - \mu)/\sigma \sim \cN(0, I)$. In this case, after one update of adversarial training, optimal model parameters will be:
\begin{align*}
    \mu^{adv} &= \bE[\bx^{adv}] \\
    & = \mu + \epsilon \bE \Big[ \frac{\tx}{\| \tx \|} \Big] = \mu \\
    K^{adv} &= \bE[(\bx^{adv} - \mu)(\bx^{adv} - \mu)^{T}] \\
        &= \sigma^2 \bE \Big[ \Big( \tx + \epsilon\frac{\tx}{\| \tx \|} \Big) \Big( \tx + \epsilon\frac{\tx}{\| \tx \|} \Big)^{T} \Big] \\
        &= \sigma^2 \Big( \bE[\tx \tx^{T}] + 2\epsilon \bE \Big[ \frac{\tx \tx^{T}}{\| \tx \|} \Big] + \epsilon^2 \bE \Big[ \frac{\tx \tx^{T}}{\| \tx \|^2} \Big]  \Big) \\
        &= \sigma^2 \Big( I + \frac{2\sqrt{2}\epsilon}{n} \frac{\Gamma((n+1)/2)}{\Gamma(n/2)} I + \frac{\epsilon^2}{n}I \Big) \\
        &= \sigma^2 I + \sigma^2 \aI = \sigma^2(1 + \alpha) I \\
        & \text{\textit{where} } \alpha = ( \frac{2\sqrt{2}\epsilon}{n} \frac{\Gamma((n+1)/2)}{\Gamma(n/2)} + \frac{\epsilon^2}{n} )
\end{align*}
The above result follows from the fact that sum of diagonal terms of $\tx \tx^{T}/\| \tx \|$ has a Chi distribution with $n$ degrees of freedom. We observe that adversarial training preserves the mean vector, but increases the variance by a multiplicative factor $1+\alpha$. 

Performing $m$ steps of adversarial training results in the following estimate:
\begin{align}\label{eq:PGD_estimate}
    \mu^{adv}_{k} &= \mu \\
    K^{adv}_{k} &=   \sigma^2(1 + \alpha)^{m}I \nonumber
\end{align}

Using the above argument, we obtain the following robustness guarantees for adversarial training:
\begin{theorem}
Let $\bx$ be an input sample drawn from $N(\mu, \sigma^2 I)$. Let $L(\bx)$ denote the log-likelihood function of the sample $\bx$ estimated using an $m$-step adversarially trained model. Let $\delta$ be any perturbation vector such that $\| \delta \| \leq \epsilon$. For any $\Delta$, when 
\begin{align}
    m \geq \max \Big[& \log \Big( \frac{1}{2\sigma^2\Delta} \Big( 2\sigma\epsilon \sqrt{20\log(1/\gamma)} + \epsilon^2 \Big) \Big),\nonumber\\
    &\log \Big( \frac{1}{2\sigma^2\Delta} \Big[ 2\sigma\epsilon\sqrt{2n} + \epsilon^2 \Big] \Big) \Big] /  \log(1+\alpha)\nonumber
\end{align}
with probability greater than $1-\gamma$,
\begin{align*}
    L(\bx) - L(\bx + \delta) < \Delta
\end{align*}
\end{theorem}
The proof for this theorem is presented in the appendix. This theorem states that, with high probability and for a sufficiently large $m$, $m$-step adversarial training learns a generative model whose likelihood estimates are provably robust within $\Delta$.

\subsubsection{Trade-off between robustness and accuracy}
The estimated parameters of our linear model after $m$ steps of adversarial training is given in Eq.~\eqref{eq:PGD_estimate}. The average $\log$-likelihood of unperturbed (clean) samples drawn from $\cN(\mu, \sigma^2 I)$ under the adversarially-trained model can be computed as
\begin{align*}
    L_{nat}(m) = &-\frac{n}{2} \log(2\pi \sigma^2 (1+\alpha)^{m}) \nonumber \\
    & - \bE_{\bx \in \cN(\mu, \sigma^2 I)} \frac{\| \bx - \mu \|^2}{2\sigma^2 (1+\alpha)^m} \nonumber \\
    = & -\frac{n}{2} \log(2\pi \sigma^2 (1+\alpha)^{m}) - \frac{n}{2(1+\alpha)^m}
\end{align*}

\begin{figure}[h]
\begin{center}
{\includegraphics[width=0.45\textwidth]{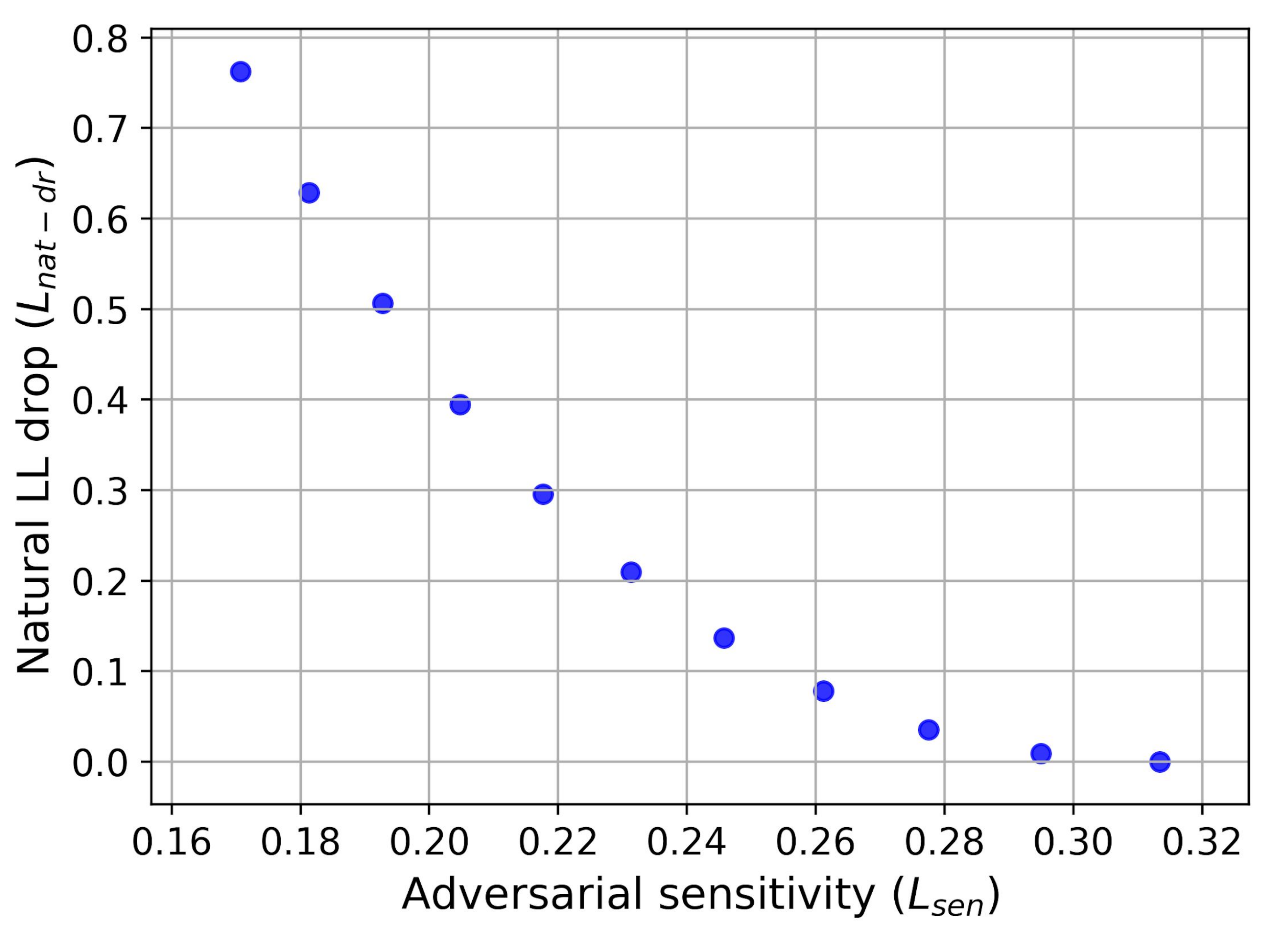}}
\end{center}
\caption{Plot showing the trade-off between performance and robustness for an example linear generative model.}
\label{fig:tradeoff}
\end{figure}

The drop in the natural likelihood due to adversarial training, which we define as $L_{nat-dr}:=L_{nat}(0) - L_{nat}(m)$, simplifies as
\begin{align*}
    L_{nat-dr}(m) = \frac{n}{2} \Big[ \log((1+\alpha)^m) + \frac{1}{(1 + \alpha)^m} - 1 \Big]
\end{align*}
$L_{nat-dr}(m)$ represents how much the average log likelihood scores will be different if we use an $m$-step adversarially-trained model instead of the clean model. Larger $m$ will lead to a larger drop in the accuracy of the likelihood computation. However, it will increase the robustness of likelihood scores against adversarial perturbations. To characterize this trade-off, note that the likelihood of perturbed samples under the $m$-step adversarially trained model can be computed as
\begin{align*}
    L_{adv}(m) = &-\frac{n}{2} \log(2\pi \sigma^2 (1+\alpha)^{m}) \nonumber \\
    & - \bE_{\bx \in \cN(\mu, \sigma^2 I)} \frac{\| \bx + \frac{\epsilon}{\| \bx - \mu \|}(\bx - \mu) - \mu\|^2}{2\sigma^2(1+\alpha)^m} \\
    = &-\frac{n}{2} \log(2\pi \sigma^2 (1+\alpha)^{m}) \\
    &- \frac{n + 2\epsilon \sqrt{2}\frac{\Gamma((n+1)/2)}{\Gamma(n/2)} + \epsilon^2}{2(1+\alpha)^m}
\end{align*}
Hence, the adversarial sensitivity, which we define as $L_{sen}(m) = L_{clean}(m) - L_{adv}(m)$ simplifies to 
\begin{align}
    L_{sen}(m) = \frac{2\epsilon \sqrt{2}\frac{\Gamma((n+1)/2)}{\Gamma(n/2)} + \epsilon^2}{2(1+\alpha)^m}
\end{align}
Adversarial sensitivity indicates the drop in likelihood scores due to adversarial attacks. Higher the score, more sensistive is the model to adversarial perturbations. We can see that $L_{nat-dr}$ is at odds with $L_{sen}$. In Figure \ref{fig:tradeoff}, we plot the trade-off between natural likelihood drop vs. the adversarial sensitivity for different values of $m$ in the range $[0, 10]$. In this experiment, we use $n=10$, and generate samples from a Gaussian distribution with a random mean and covariance matrix. We observe that the setting that gives low performance drop incurs high robustness drop, and vice-versa.

\subsection{Universal adversarial perturbation}
In universal adversarial perturbation, we are interested in finding a single perturbation vector $\delta$ such that the population likelihood (under the normal distribution) decreases maximally, i.e., we are interested in finding a perturbation $\delta$ such that 
\begin{align*}
    \min_{\delta} ~~ & C - \frac{1}{2}\log(|K|)  \\
    & -\bE_{\bx\sim \cN(\mu,K)} \left[\frac{(\bx - \mu + \delta)^T K^{-1} (\bx - \mu + \delta)}{2}\right] \\
    \text{s.t. } & \delta^{T}\delta = \epsilon^2
\end{align*}
Simplifying the objective, we obtain
\begin{align*}
    \min_{\delta} ~~ \bE_{\bx} \Big[ &-\frac{(\bx - \mu)^T K^{-1}(\bx - \mu)}{2}  \\ 
    &+ \delta^{T}K^{-1}(\bx - \mu) + \frac{\delta^T K^{-1}\delta}{2} \Big] \\
\end{align*}
The first term is independent of $\delta$, thus can be ignored. The second term is $0$ since the mean of $\bx$ is $\mu$. Thus, the optimization simplifies as 
\begin{align*}
    \max_{\delta} & \frac{\delta^{T}K^{-1}\delta}{2} \\
    \text{s.t. } & \delta^{T}\delta = \epsilon^2
\end{align*}
This is the standard Rayleigh quotient problem, the solution for which is
\begin{align*}
    \delta^* = \epsilon \lambda_{max}(K^{-1}) = \epsilon u_{min}(K) 
\end{align*}

\paragraph{Remark: } The adversarial distribution for the universal case is $\bx^{adv} = \bx + \epsilon u_{min}(K)$. The perturbation in this case only shifts the mean of the perturbed distrbution, but the covariance remains the same. Hence, adversarial training results in the following estimation: $\hat{\mu}^{adv} = \mu + \epsilon u_{min}(K)$, and $K^{adv} = K$. Since only the mean gets shifted, the resulting adversarially trained model can again be attacked with the same perturbation, resulting the same sensitivity as the clean model. Hence, adversarial training is not successful to defend against universal adversarial attacks. 

\section{Adversarial attacks and defenses on non-linear flow-based models}
\label{sens}
In this section, we empirically study the robustness of deep flow-based generative models against adversarial attacks. First, we adapt the PGD 
attack to the flow-based models by replacing the classification loss $L_{cls}$ with the $\log$-likelihood function. For normal (in-distribution) samples, we seek to compute perturbations with a bounded $\ell_{\infty}$ norm such that the likelihood of the perturbed sample is decreased maximally. 
This defines our \textit{in-distribution} attack. We also use the crafted adversarial examples to recursively re-train the model to make it more robust against adversarial attacks, we call this \textit{adversarial training}.


Next, we explore a new type of adversarial attack on the flow-based generative models where the goal of the adversary is to maximize the likelihood score of out-of-distribution (anomaly) samples to be similar to that of normal samples. We call this attack the {\it out-of-distribution adversarial attack}. This amounts to ascending (rather than descending) on the likelihood of out-of-distribution samples.
We compare these attacks to random uniform noise, which is used as a baseline.



We empirically observe that adversarially trained models obtain higher NLL (i.e. lower likelihood) on clean (un-perturbed) data than models trained on clean data alone. This is expected as no clean samples were exposed at the training time. However, this is undesirable as a good generative model should assign high likelihoods to in-distribution samples. To mitigate this problem, we propose training simultaneously on both clean and adversarial examples. In each batch of training, we mix clean and adversarial samples in 1:1 ratio. We call this procedure the \textit{hybrid adversarial training}. The analog of this method is known to fail for classification problems\citep{DBLP:journals/corr/SzegedyZSBEGF13}. However, it succeeds to robustify flow-based generative models, while preserving likelihood on unperturbed samples.

\section{Experiments}
\label{experiments}

We perform experiments on two flow-based generative models: GLOW and RealNVP, on three datasets: CIFAR-10, LSUN Bedroom, and CelebA. We evaluate the robustness of the three model varieties: models trained on clean (unperturbed) data alone, models trained on adversarially-perturbed data alone (adversarial training), and models trained on clean \textit{and} adversarial data in 1:1 ratio (hybrid training). Adversarial and hybrid models were trained with  $\epsilon=8$ and $m=10$ attack iterations. More experimental details can be found in supplementary material. In all experiments, we report negative log-likelihood values in the units of bits per dimension \citep{DBLP:journals/corr/TheisOB15}.

\begin{figure*}[h!]
\begin{center}
{\includegraphics[height=7cm]{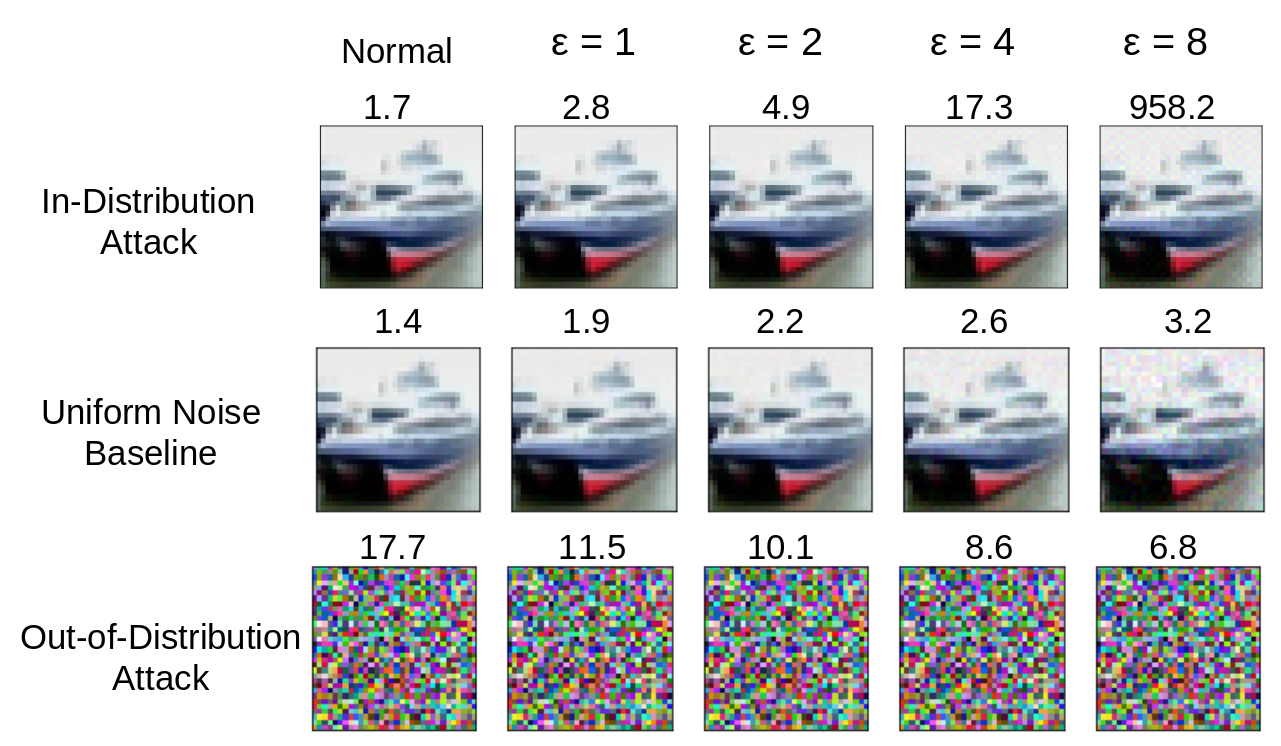}}
\end{center}
\caption{Sample visualizations of in-distribution, out-of-distribution and uniform noise attacks on a GLOW model trained on CIFAR-10 dataset. Results on clean (unperturbed) data are reported as $\epsilon=0$.}
\label{fig:cifar-10-results}
\end{figure*}

\begin{figure*}[h!]
 \begin{center}
 {\includegraphics[height=7cm]{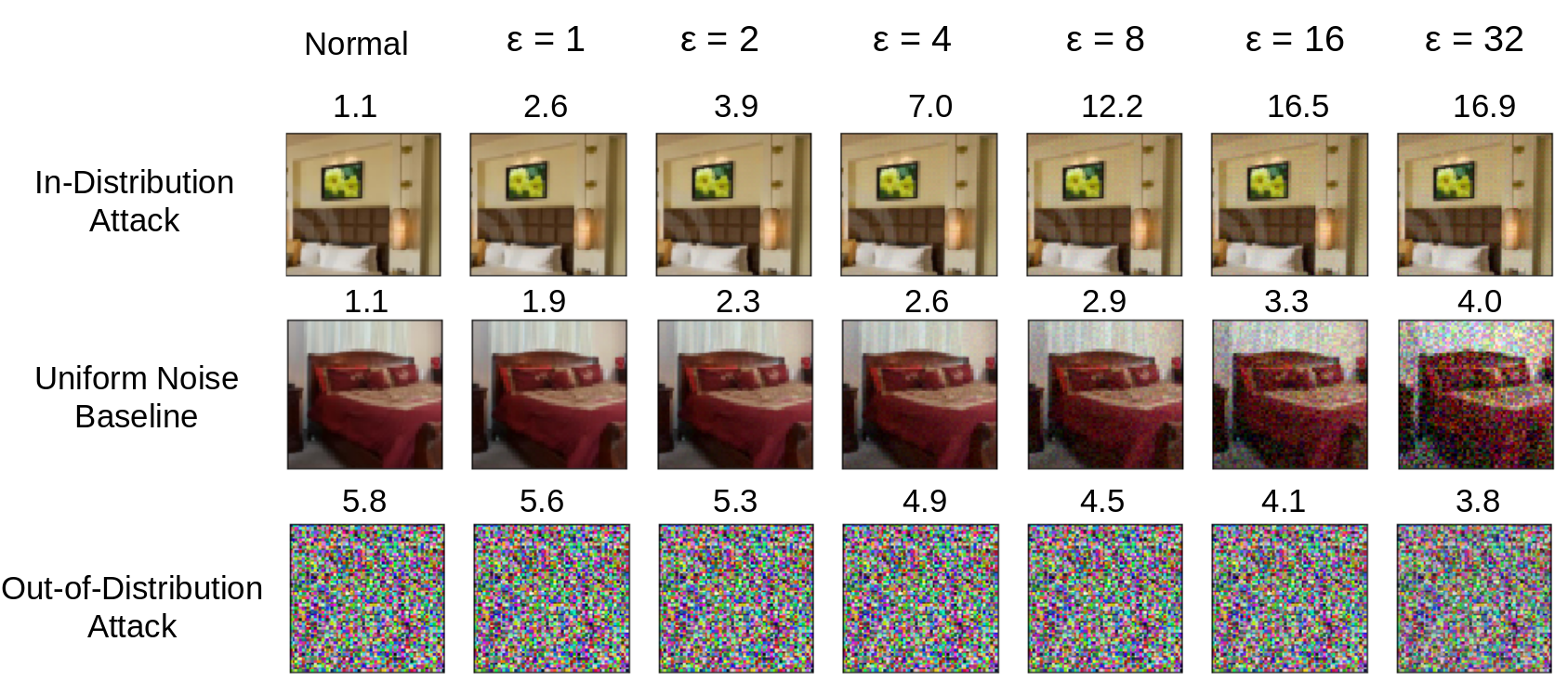}}
 \end{center}
 \caption{Sample visualizations of in-distribution, out-of-distribution and uniform noise attacks on a GLOW model trained on LSUN-Bedroom dataset. Results on clean (unperturbed) data are reported as $\epsilon=0$.}
 \label{fig:lsun-results}
 \end{figure*}

Figures \ref{fig:cifar-10-results} and \ref{fig:lsun-results} show visualizations of adversarial attacks on a GLOW model trained on unperturbed CIFAR-10 and LSUN-bedrooms datasets respectively. In the top row, we show in-distribution attacks at different attack strengths. We observe that in-distribution attacks are effective even at low $\epsilon$ values. The effectiveness of these attacks are evident as the values are much higher compared to the uniform noise baseline (shown in middle row). In the last row, we show out-of-distribution attacks, where a uniform noise image is perturbed to assign low NLL scores. For high attack strength ($\epsilon = 8$), likelihood values are on-par with values obtained by in-distribution samples.



\begin{table}[t!]
\begin{center}
\begin{tabular}{|c|c|c|c|}
\hline
\bf{$\epsilon$} & \bf{Clean} & \bf{Adv.} & \bf{Hybrid} \\\hline\hline
0 & \bf{3.4} & 4.7 & 3.6 \\
1 & 6.3 & 4.9 & \bf{4.7} \\
2 & 14 & \bf{5.0} & \bf{5.0} \\
4 & 320 & \bf{5.3} & \bf{5.3} \\
8 & 2.0$\times10^6$ & \bf{5.8} & 5.9 \\\hline
\end{tabular}
\caption{Robustness results of GLOW model trained on CIFAR-10} 
\label{tab:glow-cifar-10-robustness}
\end{center}
\end{table}

\begin{table}[t!]
\begin{center}
\begin{tabular}{|c|c|c|c|}
\hline
\bf{$\epsilon$} & \bf{Clean} & \bf{Adv.} & \bf{Hybrid} \\\hline\hline
0 & \bf{2.4} & 4.4 & 2.9 \\
1 & 5.5 & \bf{4.5} & 4.7 \\
2 & 8.5 & 4.7 & \bf{4.6} \\
4 & 15.2 & \bf{5.0} & \bf{5.0} \\
8 & 27.0 & \bf{5.5} & 5.6 \\
16 & 35.8 & \bf{6.6} & \bf{6.6} \\
32 & 36.4 & \bf{7.7} & 8.1 \\\hline
\end{tabular}
\caption{Robustness results of GLOW model trained on LSUN-Bedrooms dataset} 
\label{tab:glow-lsun-bedroom-robustness}
\end{center}
\end{table}

\begin{table}[t!]
\begin{center}
\begin{tabular}{|c|c|c|}
\hline
\textbf{Attack Iterations} & \textbf{CelebA} & \textbf{LSUN} \\
\hline\hline
0 & 2.9 & 2.7 \\ 
10 & 14.3 & 10.9 \\
20 & 17.9 & 14.4 \\
50 & 24.9 & 19.8 \\ \hline
Uniform noise & 5.8 & 4.8 \\ \hline
\end{tabular}
\caption{Adversarial attacks on RealNVP models trained on CelebA and LSUN Bedroom. All models were attacked with $\epsilon=8$. }
\label{tab:realNVP-celebA-robustness}
\end{center}
\end{table}


Next, we present quantitative results, where we report average sample likelihood scores (in bits/per dimension), averaged over the test set. Likelihood scores on adversarial samples over a sweep of attack strength (attack $\epsilon$) for a GLOW model trained on CIFAR-10 and LSUN Bedroom datasets are shown in Tables \ref{tab:glow-cifar-10-robustness} and \ref{tab:glow-lsun-bedroom-robustness} respectively. We observe that adversarially trained models improve robustness, however the NLL scores on unpertubed samples increase drastically. Hybrid adversarial training, on the other hand, achieves (1) NLL on adversarial samples comparable to adversarially trained model, and (2) likelihood on unperturbed samples comparable to clean baseline, i.e., hybrid model improves robustness preserving the performance on unperturbed samples, thus achieving the best of both worlds.  

In Table \ref{tab:realNVP-celebA-robustness}, we report robustness results for RealNVP trained on CelebA and LSUN Bedroom datasets over a sweep of attack iterations. Due to computational constraints, we use a fixed $\epsilon=8$. The results show that RealNVP is also susceptible to adversarial attacks, similar to GLOW models.



In addition to these, we show some interesting results where samples generated from a model trained on perturbed images have an adversarial effect on a model trained on unpertubed samples. This experiment can be found in supplementary material.

\section{Conclusion}
In this paper, we present a comprehensive analysis of adversarial robustness of flow-based generative models. First, we a perform a sensitivity analysis of linear generative models, and show that adversarial training provably improves robustness. Then, we demonstrate adversarial attacks on two non-linear flow-based generative models - GLOW and RealNVP. To improve the robustness of these models, we investigate the use of adversarial training, a popular defense mechanism used in classification. We show that adversarial training improves robustness at the cost of decrease in likelihood on unperturbed data. To remedy this issue, we propose \textit{hybrid adversarial training}, a novel defense mechanism that improves adversarial robustness with a marginal drop in likelihood on unperturbed data.

\section{Acknowledgements}
This work was supported in part by NSF award CDS\&E:1854532 and award HR00111990077.

\newpage

\bibliography{references}
\bibliographystyle{bib_style}

\newpage

\section{Appendix}

\subsection{Proof of Theorem 3.1}

\begin{theorem}\label{thm:pert_gen}
  Let $L(\bx)$ denote the likelihood function of an input sample $\bx$ under a Gaussian distribution $\cN(\mu, K)$. Let $K = U \Lambda U^T$ be the Eigen-decomposition of the covariance matrix $K$. Let $c = [c_1, c_2, \hdots, c_n] = U^T K$, and $\Lambda = diag([\lambda_1, \hdots \lambda_n])$. Let $\eta$ be a solution of the set of equations
    \begin{align*}
    \sum_{i} \frac{c_i^{2}}{(1 - 2\eta \lambda_i)^{2}} &= \epsilon^2 \\
    2\eta\lambda_i - 1 &\geq 0 \quad \forall i
    \end{align*}
  Then, the optimal additive perturbation $\delta$ with norm bound $\| \delta \|_{2} < \epsilon$ that maximally decreases the likelihood score of sample $\bx$ is given by
  \begin{align}\label{eq:pert_soln}
      \delta^* = (K^{-1} - 2\eta I)^{-1} K^{-T} (\mu - \bx)
  \end{align}
\end{theorem}
\paragraph{Proof: }
We are interested in generating an adversarial attack on linear models trained on Gaussian input distribution. As explained in Section 2.1 of the main paper, adversarial perturbation $\delta$ on sample $\bx$ can be obtained by solving the following optimization: 
\begin{align*}
    \min_{\delta} \quad & C - \frac{1}{2}\log(|K|) - \frac{(\bx - \mu + \delta)^{T}K^{-1}(\bx - \mu + \delta)}{2} \\
    \text{s.t. } & \| \delta \|_{2} < \epsilon
\end{align*}
The Lagrangian $L$ corresponding to this optimization can be written as
\begin{align*}
    L =& -\frac{(\bx - \mu + \delta)^{T}K^{-1}(\bx - \mu + \delta)}{2} + \eta(\delta^{T}\delta - \epsilon^2) \\
      =& -\frac{(\bx - \mu )^{T}K^{-1}(\bx - \mu)}{2} - (\bx - \mu)^{T}K^{-1}\delta  \\
      & - \frac{\delta^{T}K^{-1}\delta}{2} + \eta(\delta^{T}\delta - \epsilon^2)
\end{align*}
\paragraph{First-order necessary conditions (KKT)}
From the stationarity condition of KKT, the gradient of the Lagrangian function w.r.t the optimization variables should be $0$.
\begin{align}\label{eq:lagran1}
    \nabla_{\delta}L &= -K^{-T}(\bx - \mu) - K^{-1}\delta + 2\eta\delta = 0 \nonumber \\
    (K^{-1} - 2\eta I)\delta &= K^{-T}(\mu - \bx) \nonumber \\
    \delta &= (K^{-1} - 2\eta I)^{-1} K^{-T} (\mu - \bx)
\end{align}
From complementary slackness, we obtain
\begin{align}\label{eq:lagran2}
    \eta( \delta^{T}\delta - \epsilon^{2} ) = 0
\end{align}
So, either $\eta=0$ or $\delta^{T} \delta = \epsilon^2$. When $\eta=0$, $\delta=\mu - \bx$. For the other condition $\delta^{T} \delta = \epsilon^2$, we obtain,
\begin{align*}
    (\mu-\bx)^{T}K^{-1}(K^{-1} - 2\eta I)^{-2}K^{-T}(\mu - \bx) = \epsilon^{2}
\end{align*}
Now, consider the Eigen-decomposition of matrix $K = U\Lambda U^{T}$, where $\Lambda = diag(\lambda_1, \lambda_2, \hdots \lambda_n)$. Using this in the above equation, we obtain the condition
\begin{align}\label{eq:condition_eta}
    \sum_{i} \frac{c_i^{2}}{(1 - 2\eta \lambda_i)^{2}} = \epsilon^2
\end{align}
where $c = [c_1, c_2, \hdots c_n] = U^{T}(\mu - \bx)$. Eq.~\eqref{eq:condition_eta} can be solved numerically to obtain the value of $\eta$.
\paragraph{Second order sufficiency condition:}
The Hessian of the Lagrangian function can be written as
\begin{align*}
    \nabla^{2}_{\delta \delta} L = -K^{-1} + 2\eta I
\end{align*}
The above matrix should be positive semi-definite. This gives the following condition
\begin{align}\label{eq:socs}
   2\eta\lambda_i - 1 &\geq 0 \quad \forall i 
\end{align}
We see that the solution $\delta=0$ does not satisfy this property, hence, it can be eliminated. Hence, the optimal perturbation is the solution to Eq.~\eqref{eq:condition_eta} which satisfy Eq.~\eqref{eq:socs}.

\subsection{Proof of Theorem 3.2}

\begin{lemma}\label{lem:chi-square-tailbound}
Let $X$ be a  $\chi^2(n)$ distribution. Then, for any $t>1$,
\begin{align*}
    Pr(X \geq 2tn) \leq e^{-\frac{tn}{10}}
\end{align*}
\end{lemma}
\paragraph{Proof: } From \cite{laurent2000adaptive}, we know that for a $\chi^2(n)$ random variable $X$,
\begin{align*}
    Pr(X \geq n + 2\sqrt{nx} + 2x) \leq e^{-x}
\end{align*}
Substituting $x = \frac{tn}{10}$, we get
\begin{align*}
    Pr(X \geq n + 2n\sqrt{t/10} + 2tn/10) \leq e^{-tn/10}
\end{align*}
Now, $n(1 + 2\sqrt{t/10} + 2t/10) < n(1 + 2t/10 + 2t/10) < 2nt$ for $t>1$. Hence, 
\begin{align*}
    Pr(X \geq 2nt) & < Pr(X \geq n + 2n\sqrt{t/10} + 2tn/10) \\
    & \leq e^{-tn/10}
\end{align*}

\begin{theorem}
Let $\bx$ be an input sample drawn from $N(\mu, \sigma^2 I)$. Let $L(\bx)$ denote the log-likelihood function of the sample $\bx$ estimated using a $m$- step adversarially trained model. Let $\delta$ be any perturbation vector having a norm-bound $\| \delta \|_2 \leq \epsilon$. For any $\Delta$, when $m \geq max \Big[ \log \Big( \frac{1}{2\sigma^2\Delta} \Big( 2\sigma\epsilon \sqrt{20\log(1/\gamma)} + \epsilon^2 \Big) \Big) ,  \log \Big( \frac{1}{2\sigma^2\Delta} \Big[ 2\sigma\epsilon\sqrt{2n} + \epsilon^2 \Big] \Big) \Big] /  \log(1+\alpha)$, with probability greater than $1-\gamma$,
\begin{align*}
    L(\bx) - L(\bx + \delta) < \Delta
\end{align*}
\end{theorem}
\paragraph{Proof: }
From Section 3.1.1 of main paper, the optimal adversarial perturbation for spherical covariance matrix is given by
\begin{align*}
    \delta = \frac{\epsilon}{\| \bx - \mu \|} (\bx - \mu)
\end{align*}{}
From Section 3.2, we know that the estimated model parameters after $m$ steps of adversarial training is given by
\begin{align*}
    \mu_{m}^{adv} &= \mu \\
    K_{m}^{adv} &= \sigma^2 (1 + \alpha)^m I
\end{align*}
Log-likelihood drop for this model under the optimal adversarial perturbation can then be computed as
\begin{align*}
    L(\bx) &= C' - \frac{\| \bx - \mu \|^2}{2\sigma^2(1+\alpha)^m} \\
    L(\bx + \delta) &= C' - \frac{\| \bx - \mu \|^2}{2\sigma^2(1+\alpha)^m} \Big(1 + \frac{\epsilon}{\| \bx - \mu \|} \Big)^2 \\
    L(\bx) - L(\bx + \delta) &= \frac{1}{2\sigma^2(1+\alpha)^m} \Big( 2\epsilon\| \bx - \mu \| + \epsilon^2\Big)
\end{align*}

We want this likelihood difference to be less than $\Delta$.
\begin{align*}
    & Pr(L(\bx) - L(\bx + \delta) < \Delta) \\
    & = Pr(\frac{\| \bx - \mu \|}{\sigma}  < \frac{2\sigma^2\Delta(1+\alpha)^m - \epsilon^2}{2\sigma\epsilon})
\end{align*}
We now reparameterize $\tx = \frac{\bx - \mu}{\sigma} \sim \cN(0, I)$. The norm vector $\| \tx \|^2$ then obeys a $\chi^2(n)$ distribution with $n$ degrees of freedom. Then, 
\begin{align}\label{eq:chi-square}
    & Pr(L(\bx) - L(\bx + \delta) < \Delta) \nonumber \\
    & = Pr\Big( \| \tx \|^2 < \Big( \frac{2\sigma^2\Delta(1+\alpha)^{m} - \epsilon^2}{2\sigma\epsilon} \Big)^2 \Big) \\
        &= 1 - Pr\Big( \| \tx \|^2 \geq \Big( \frac{2\sigma^2\Delta(1+\alpha)^{m} - \epsilon^2}{2\sigma\epsilon} \Big)^2 \Big) \nonumber
\end{align}
Now, we use Lemma~\ref{lem:chi-square-tailbound} in \eqref{eq:chi-square}. Set \begin{align}\label{eq:t-cond}
    t = \frac{1}{2n} \Big( \frac{2\sigma^2\Delta(1+\alpha)^{m} - \epsilon^2}{2\sigma\epsilon} \Big)^2
\end{align}
Then, 
\begin{align*}
    & Pr\Big( \| \tx \|^2 \geq \Big( \frac{2\sigma^2\Delta(1+\alpha)^{m} - \epsilon^2}{2\sigma\epsilon} \Big)^2 \Big) \\ 
    & \leq \exp{ \Big( \frac{-1}{20} \Big( \frac{2\sigma^2\Delta(1+\alpha)^{m} - \epsilon^2}{2\sigma\epsilon} \Big)^2} \Big) \\
    & \leq \gamma
\end{align*}
Simplifying the above expression, we obtain,
\begin{align*}
    m \geq \frac{\log \Big( \frac{1}{2\sigma^2\Delta} \Big( 2\sigma\epsilon \sqrt{20\log(1/\gamma)} + \epsilon^2 \Big) \Big) }{\log(1 + \alpha)}
\end{align*}
Also, in condition \eqref{eq:t-cond}, we require $t>1$. This gives, 
\begin{align*}
m > \frac{ \log \Big( \frac{1}{2\sigma^2\Delta} \Big[ 2\sigma\epsilon\sqrt{2n} + \epsilon^2 \Big] \Big) }{\log(1+\alpha)}
\end{align*}

Hence, when $m \geq max\Big[ \log \Big( \frac{1}{2\sigma^2\Delta} \Big( 2\sigma\epsilon \sqrt{20\log(1/\gamma)} + \epsilon^2 \Big) \Big) ,  \log \Big( \frac{1}{2\sigma^2\Delta} \Big[ 2\sigma\epsilon\sqrt{2n} + \epsilon^2 \Big] \Big) \Big] /  \log(1+\alpha)$, $Pr(L(\bx) - L(\bx + \delta) < \Delta) \geq 1-\gamma$
. This concludes the proof.


\begin{figure*}[h]
 \begin{center}
 {\includegraphics[height=5cm]{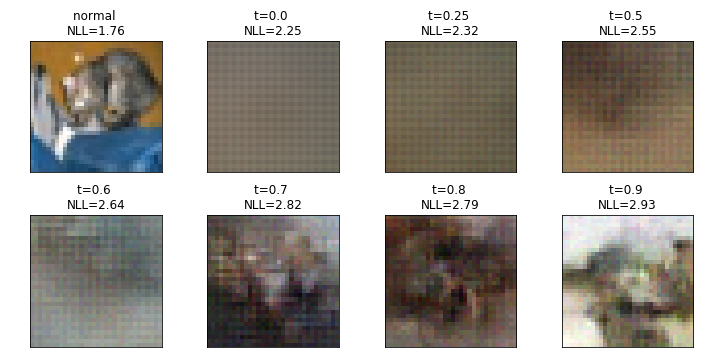}}
 \end{center}
 \caption{ Samples generated at different temperatures from an adversarially trained model, evaluated against a clean model.}
 \label{fig:gen_samples_are_adv}
 \end{figure*}

\subsection{Are generated samples from adversarially trained model adversarial?}

Generative models trained on adversarial examples provide a unique opportunity to ask the question of whether the samples generated by this model has an adversarial nature. To do this, we generate samples from an adversarially trained model, and evaluate their likelihood on model trained on unperturbed samples. We find that samples generated by adversarially trained model are indeed adversarial with respect to the unperturbed model, at a strength comparable to that on which the model was trained. These results are shown in in Figure \ref{fig:gen_samples_are_adv}.

%

\subsection{Experimental Details}

All model architectures for GLOW and RealNVP were trained with default values given in their respective implementations. Adversarial and Hybrid models were trained with $\epsilon=8$, and $m=10$ attack iterations. GLOW test sizes for CIFAR-10 and LSUN Bedroom test size were $N=10,000$ and $N=1200$ (default) respectively. For GLOW robustness evaluations, adversaries were trained with $m=32$ and $m=40$ for CIFAR-10 and LSUN Bedroom respectively. 

For the random noise baseline, random images were generated as $\text{Unif}[-\epsilon,\epsilon]$ (centered) and then clipped to $[0,255]$.

Out-of-distribution attacks were performed with a $\text{Unif[0,255]}$ random image, and trained trained with $m=100$ iterations.

\subsection{Instability in GLOW likelihood evaluations on CIFAR-10 for high $\epsilon$}

In our experiments, we observed variance in likelihood evaluations for GLOW models trained on CIFAR-10 under strong adversaries (attack strengths $\epsilon \geq 8$). On the other hand, low attack strengths ($\epsilon < 8$) had negligible variance. This adds uncertainty as to the "true" value of the attack strength, however we maintain that the trend is clear: stronger adversaries are more disruptive of the likelihood score. As this paper is primarily concerned with the existence of adversarial attacks and robust defenses, we consider this issue not germane to the present work. For completeness, we give details on our investigation of this issue below.

Sources of randomness and numerical issues were investigated. Two sources of randomness found were (1) the addition of uniform random noise in the computation of \textit{continuous} log-likelihoods (Equation (2) in \citet{NIPS2018_8224}) and (2) the random initialization of rotation matrices $\mathbf{W}$ in the invertible $1\times1$ convolution (paragraph below Equation (9) in \citet{NIPS2018_8224}).

High NLL values correspond to very small probabilities. Since the entire computation is done in the $\log$ scale, underflow is not a problem. Computations are by default performed with \lstinline{float32}, having the range of approximately $\pm 3.4 \times 10^{38}$, which far exceeds the highest value we observed of $10^{16}$.

The authors of \citet{NIPS2018_8224} propose LU-decomposition as a fast means of computing the determinant. In the reference implementation this option was disabled by default. We found that enabling it helped with numeric stability, with negligible drop in training speed.



\end{document}